\documentclass[a4paper]{article}

\usepackage{arxiv}

\usepackage[utf8]{inputenc} 
\usepackage[T1]{fontenc}    
\usepackage[breaklinks]{hyperref}
\usepackage{subfigure}
\usepackage{url}
\usepackage{breakurl}
\usepackage{booktabs}       
\usepackage{amsfonts}       
\usepackage{nicefrac}       
\usepackage{microtype}      
\usepackage{amsmath}
\usepackage{amsfonts}
\usepackage{cite}
\usepackage{amssymb}
\usepackage{amsthm}
\usepackage{verbatim}
\usepackage{enumerate}
\usepackage{mdwlist} 
\usepackage{color}
\usepackage[utf8]{inputenc}

\usepackage{algpseudocode}
\usepackage{algorithmicx}
\usepackage{lineno}
\usepackage{framed,multirow}
\usepackage[final]{pdfpages}
\usepackage{latexsym}
\usepackage{wrapfig}

\usepackage{verbatim}
\usepackage{bm}
\usepackage{authblk}

\expandafter\def\expandafter\UrlBreaks\expandafter{\UrlBreaks
    \do\a\do\b\do\c\do\d\do\e\do\f\do\g\do\h\do\i\do\j%
    \do\k\do\l\do\m\do\n\do\o\do\p\do\q\do\r\do\s\do\t%
    \do\u\do\v\do\w\do\x\do\y\do\z\do\A\do\B\do\C\do\D%
    \do\E\do\F\do\G\do\H\do\I\do\J\do\K\do\L\do\M\do\N%
    \do\O\do\P\do\Q\do\R\do\S\do\T\do\U\do\V\do\W\do\X%
    \do\Y\do\Z\do\/\do-}

\title{Learning Local Complex Features using Randomized Neural Networks for Texture Analysis}

\author[1]{Lucas C. Ribas}
\author[2]{Leonardo F. S. Scabini}
\author[3]{Jarbas Joaci de Mesquita S\'{a} Junior}
\author[1,2]{Odemir M. Bruno}
\affil[1]{\small{S\~{a}o Carlos Institute of Physics, University of S\~{a}o Paulo (USP), PO Box 369, 13560-970, S\~{a}o Carlos, SP, Brazil. \protect\\Scientific Computing Group}}
\affil[2]{\small{Institute of Mathematics and Computer Science, University of S\~{a}o Paulo (USP), USP, Avenida Trabalhador s\~ao-carlense, 400, 13566-590, S\~ao Carlos, SP, Brazil.}}
\affil[3]{\small{Curso de Engenharia da Computa\c{c}\~ao,
Programa de P\'os-Grad. em Eng. El\'etrica e de Computa\c{c}\~ao, 
Campus de Sobral, Universidade Federal do Cear\'a,
Rua Coronel Estanislau Frota, 563, Centro,
Sobral, Cear\'a, CEP: 62010-560, Brasil}}

\begin{document}
\maketitle

\begin{abstract}
Texture is a visual attribute largely used in many problems of image analysis. Currently, many methods that use learning techniques have been proposed for texture discrimination, achieving improved performance over previous handcrafted methods.
In this paper, we present a new approach that combines a learning technique and the Complex Network (CN) theory for texture analysis.
This method takes advantage of the representation capacity of CN to model a texture image as a directed network and uses the topological information of vertices to train a randomized neural network.
This neural network has a single hidden layer and uses a fast learning algorithm, which is able to learn local CN patterns for texture 
characterization. Thus, we use the weighs of the trained neural network to compose a feature vector.
These feature vectors are evaluated in a classification experiment in four widely used image databases. 
Experimental results show a high classification performance of the proposed method when compared to other methods, indicating that our approach can be used in many image analysis problems.
\end{abstract}


\section{Introduction}
Texture is a key visual feature used to describe images in many problems of computer vision and image processing such as plant recognition, medical image analysis, industrial inspection, and microscope images.
The visual texture is composed of sub-patterns related to the pixel distribution and with certain similarity on the image \cite{gonccalves2016texture}.
Although it can be easily understood by the humans, there is no formal definition for the texture attribute.
However, this fact did not prevent the progress and development of new approaches for texture analysis.

Many techniques for texture analysis have been proposed in the last decades. They can be divided into four classical categories according to the way of exploiting the texture characteristics of the image.
The statistical-based approaches have been the most studied over the last decades.
Examples of these approaches are the variants of Gray-Level Co-occurrence Matrices (GLCM) \cite{haralick1973} and Local Binary Patterns (LBP) \cite{ojala2002multiresolution}.
The structural approaches treat the image as an arrangement of textons (ie. small elements), which form the texture from a spatially organized pattern. 
Some methods of this kind of analysis are the Morphological decomposition \cite{lam1997rotated} and key point detectors and descriptors to represent the texture elements \cite{lazebnik2005sparse}.  
On the other hand, in the spectral-based approaches, the image is exploited in the power spectrum domain and the most popular methods are the Gabor filters \cite{manjunath1996texture} and Wavelet transforms \cite{DEVES20142925}.
Finally, in the model-based approaches, the textures are represented using mathematical models and estimation of their parameters such as Fractal models \cite{backes2012,Ribas2015} and stochastic models \cite{panjwani1995markov}.

More recently, innovative methods have been proposed to characterize textures, achieving promising results.
In particular, methods that use learning techniques to represent the texture have gained prominence such as that ones based on a vocabulary of Scale Invariant Feature Transform (SIFT) \cite{2004densesift} (called Bag-Of-Visual-Words (BOVW)), randomized neural networks \cite{JarbasRNN2015,SaJrRNNColor2019} and deep convolutional neural networks \cite{simonyan2014very,szegedy2016rethinking,szegedy2017inception,basu2016theoretical}.
On the other hand, methods based on image complexity analysis have also gained attention due to its capacity to deal with the complex texture patterns. In this sense, one of the most popular are the methods based on Complex Network (CN) theory. These methods are very promising because of its ability to represent the relationships among structural elements of texture.
However, once the textures are modeled as CNs,  how to characterize it in order to obtain representative descriptors is a challenge to overcome.


In this paper, we propose to use a randomized neural network to learn and characterize the topology of a directed CN that models a texture image.
Firstly, the image is modeled as a directed CN mapping the pixels into vertices and connecting the vertices based on a connection rule.
For characterization, we use each vertex of the CN as a label and the neighboring vertices (ie. the eight neighboring pixels in a window $3\times3$) as an input vector for training a randomized neural network.
A randomized neural network has a single-hidden layer
with a very fast learning algorithm that can learn and characterize the topological characteristics of the CN.
Thus, the feature vector of texture is composed of the output weights of a trained randomized neural network.
In relation to the previous approach \cite{ribas2018fusion}, the main contribution of this approach is a new way to build the label and input vectors to train the randomized neural network. The proposed approach is faster and improves the classification performance.

This paper is organized as follows. Section \ref{sec:background} describes the fundamentals of complex network theory and randomized neural network. The proposed method for texture analysis is described in Section \ref{sec:method}. Experimental setup used in this work is presented in Section \ref{sec:experiments}, The experimental results, discussion and comparison on four databases are presented in Section \ref{sec:results}. Finally, Section \ref{sec:conclusion} concludes the paper and suggests future works.

 

\section{Background} \label{sec:background}

\subsection{Complex Networks} 

The Complex Network (CN) research, also called Network Science, arises from the combination between the graph theory, physics and statistics, targeting large and complicated real systems. Basically, most of these systems can be modeled as a structure composed of elements that interact with each other, which in a network are described by vertices and edges. However, the mathematical properties that governs these networks internal behavior is not trivial, as an example one can imagine the structural organization and functioning of the internet, composed of countless routers and computers (vertices) and their physical connections through wires (edges). These systems are collectively called complex systems, capturing the fact that it is difficult to derive their collective behavior from a knowledge of the system's components \cite{barabasi2016network}. Initially, it was believed that most of the real networks had random topology \cite{randomCN,randomCNevolution}. Further research then has shown various structural patterns, which led to the definition of network models such as the scale-free \cite{scalefreeCN} (power-law-like degree distribution) and the small-world \cite{smallworldCN} (short average path distance and high vertex inter-connectivity). These phenomena were observed in several real systems, which then allowed important advances in the study of their functioning. Therefore, the complex network approach became popular on the analysis of various real systems in areas such as physics, biology, sociology and many more. 


A Complex Network is usually defined by a combination of two sets, its vertices and edges. Let us define $V=\{v_1,...,v_n\}$ as a set of $n$ vertices and $E=\{e_{v_i, v_j}\}$ as a set of $m$ edges connecting vertex pairs, then a network is defined by $N=(V,E)$. Here we will focus on directed weighted networks, which implies that $e_{v_i, v_j} \neq e_{v_j,v_i}$ and $e_{v_i, v_j} \in \mathbb{R}$. To analyze a system from a network perspective, the first step is the modelling, which means defining the vertices and edges. Once the network $N$ is built, several topological measures can be obtained to describe its structure \cite{costa2007CNsurvey}. One of the most traditional measures is the vertex degree, that counts the number of connections of a vertex

\begin{equation}\label{eq:out_degree}
    k_{v_{i}} = \sum_{\forall v_j \in V}\left\{
    \begin{array}{l}
    1 $, \ if $e_{v_i, v_j} \in E
    \\
    0 $,  \ otherwise$
    \end{array}
    \right.
    \end{equation}

 In a directed network scenario, Equation \ref{eq:out_degree} describes the output degree of vertex $v_i$, i.e. the number of connections leaving $v_i$. It is also possible to compute the input degree by inverting the verification from "if $e_{v_i, v_j} \in E$" to "if $e_{v_j, v_i} \in E$". The traditional degree measure does not take into consideration the edge weights, therefore it is possible to calculate the weighted degree, also known as strength of a vertex
 
 \begin{equation}\label{eq:out_strength}
    ks_{v_{i}} = \sum_{\forall v_j \in V}\left\{
    \begin{array}{l}
    e_{v_i, v_j} $, \ if $e_{v_i, v_j} \in E
    \\
    0 $, \ otherwise$
    \end{array}
    \right
    .
    \end{equation}
which can also be computed for input connections, which we will denote $ke_{v_{i}}$. From the degree and strength of the vertices, many network properties can be quantified like its wiring patterns, the existence of the scale-free phenomenon, the identification of hubs and influent vertices, etc.

\subsection{Randomized Neural Network}

Randomized neural networks \cite{schmidt1992feedforward,pao1992functional,pao1994learning,huang2006extreme} are artificial neural nets, which, in their simplest version, have a single hidden layer, whose weights are determined randomly, and an output layer whose weights can be determined using a closed-form solution. When these neural networks allow direct links between the input feature vectors and the output layer \cite{pao1992functional,pao1994learning}, they are known as random vector functional link (RVFL) nets.

To mathematically explain a randomized neural network without direct links, let $X=\left[\vec{x_1}, \vec{x_2},\ldots, \vec{x_N}\right]$ be a set of $N$ input feature vectors, each one having $p+1$ attributes (each vector has an additional fixed value $+1$ for bias). Next, these input vectors can be processed by the hidden neurons by using $Z=\phi(WX)$, where $\phi(.)$ is a transfer function and $W$ is a matrix of weights of the hidden neurons whose dimensions are $Q\times(p+1)$, where $Q$ is the number of neurons of the hidden layer.

The matrix $Z=\left[\vec{z_i}, \vec{z_2},\ldots, \vec{z_N}\right]$ represents the outputs of the hidden layer for all input feature vectors, that is, $\vec{x_i} \rightarrow \vec{z_i}$. This matrix, after the inclusion of an additional fixed value $+1$ for bias in each vector, can be used to compute the weights of the output layer, according to 

\begin{equation}
M=DZ^{T}(ZZ^{T})^{-1},
\label{eq:rnn}
\end{equation}
where $D=\left[\vec{d_i}, \vec{d_2},\ldots, \vec{d_N}\right]$ is a matrix of labels (each $\vec{d_i}$ corresponding to its respective input vector $\vec{x_i}$) and $Z^{T}(ZZ^{T})^{-1}$ is the Moore-Penrose pseudo-inverse \cite{Moore1920,penrose_1955}.

Sometimes, the square matrix $ZZ^{T}$ is near-singular, resulting in unstable results for the matrix $M$. To solve this, it is possible to use the Tikhonov regularization \cite{tikhonov1963,calvetti2000}, as follows

\begin{equation}
M=DZ^{T}(ZZ^{T}+\lambda I)^{-1},
\label{eq:rnn_tik}
\end{equation}
where $I$ is an identity matrix $(Q+1)\times(Q+1)$ and $\lambda$ is a regularization parameter ($0<\lambda<1$). 

\section{Proposed Method} \label{sec:method}

In this section, we describe the proposed method in detail, from the image modeling as a directed complex network to the characterization through learning local topological properties with randomized neural networks.

\subsection{Modeling Texture as Directed Networks} 

Most of previous works approaching texture as Complex Networks considers each image pixel as a network vertex, and connections are based on their intensity similarity and spatial proximity. This is achieved through two modeling parameters: a radius, which defines a distance limit for connections, and a threshold for connection cutting. On the other hand, we employ a directed modeling as introduced in \cite{ribas2018fusion}, where only the radius is needed, removing the need of finding ideal threshold values. Consider $I$ as the input image with pixels $I(i) \in [0, L]$, where $L$ defines the maximum intensity value of pixels, usually $255$ for 8-bit images. A network $N^r(V,E)$ is built where $V=\{v_i\}$  $\forall$ $i \in I$ represent vertices, which maps each image pixel, and $E=\{e_{v_i, v_j}\}$ represent edges, connecting pairs of vertices. To create directed connections each vertex/pixel is centered in a sliding window of radius $r$ and edges points in the direction of the gradient, i.e. towards pixels of higher intensity

\begin{equation}
E = \left \{e_{v_i,v_j} \in E \mid d(v_i,v_j) \leqslant r   \wedge  I(i) < I(j) \right \}
\end{equation}
where $d(v_i,v_j)$ represents the Euclidean distance between pixels $i$ and $j$. If the pixel intensity is equal, the edge is bidirectional.

\begin{figure*}[!ht]
	\centering
    \subfigure[Directed network structure for $R=\{1,2,3\}$.]{\includegraphics[width=0.25\textwidth]{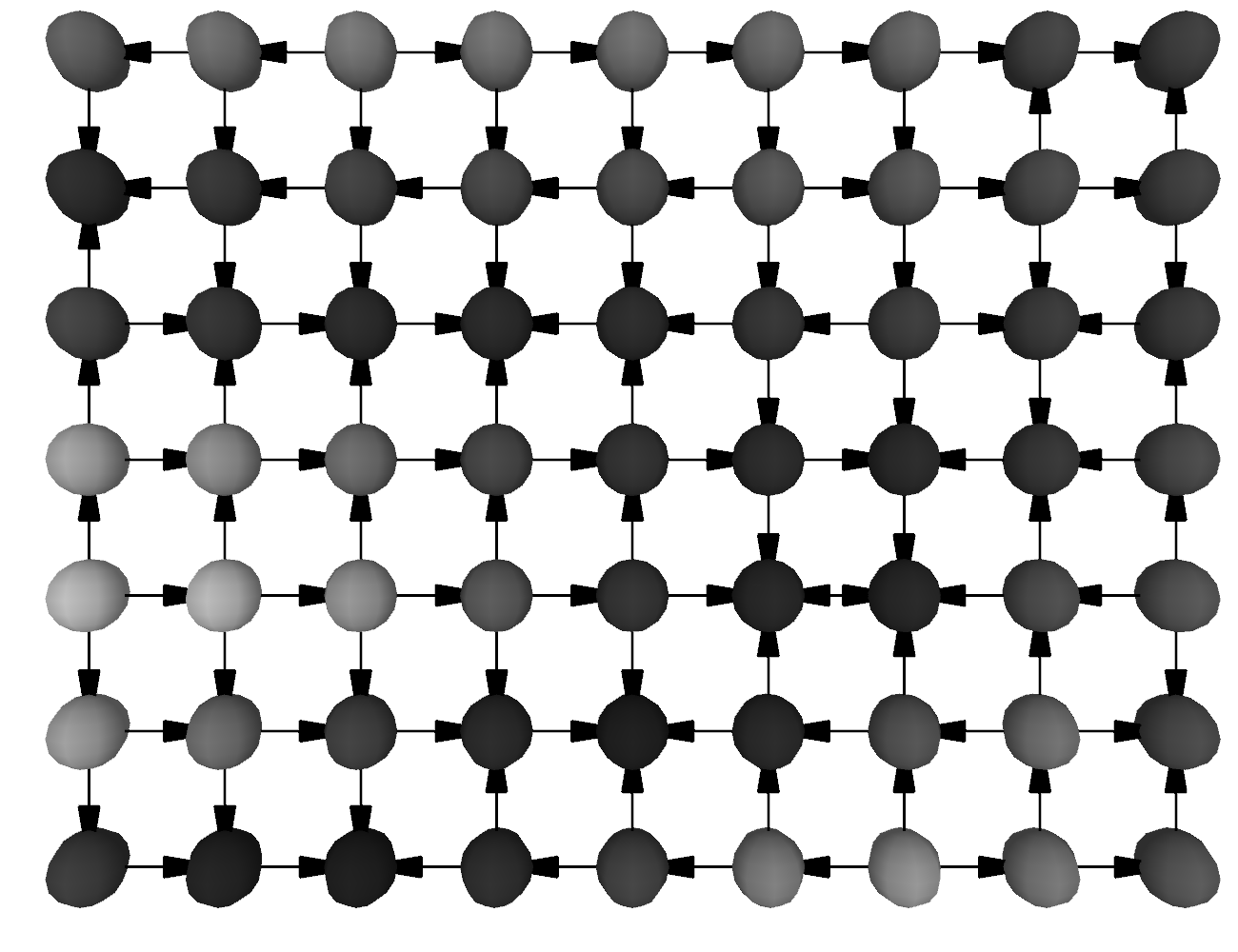} \ \ \ \ \  \includegraphics[width=0.22\textwidth]{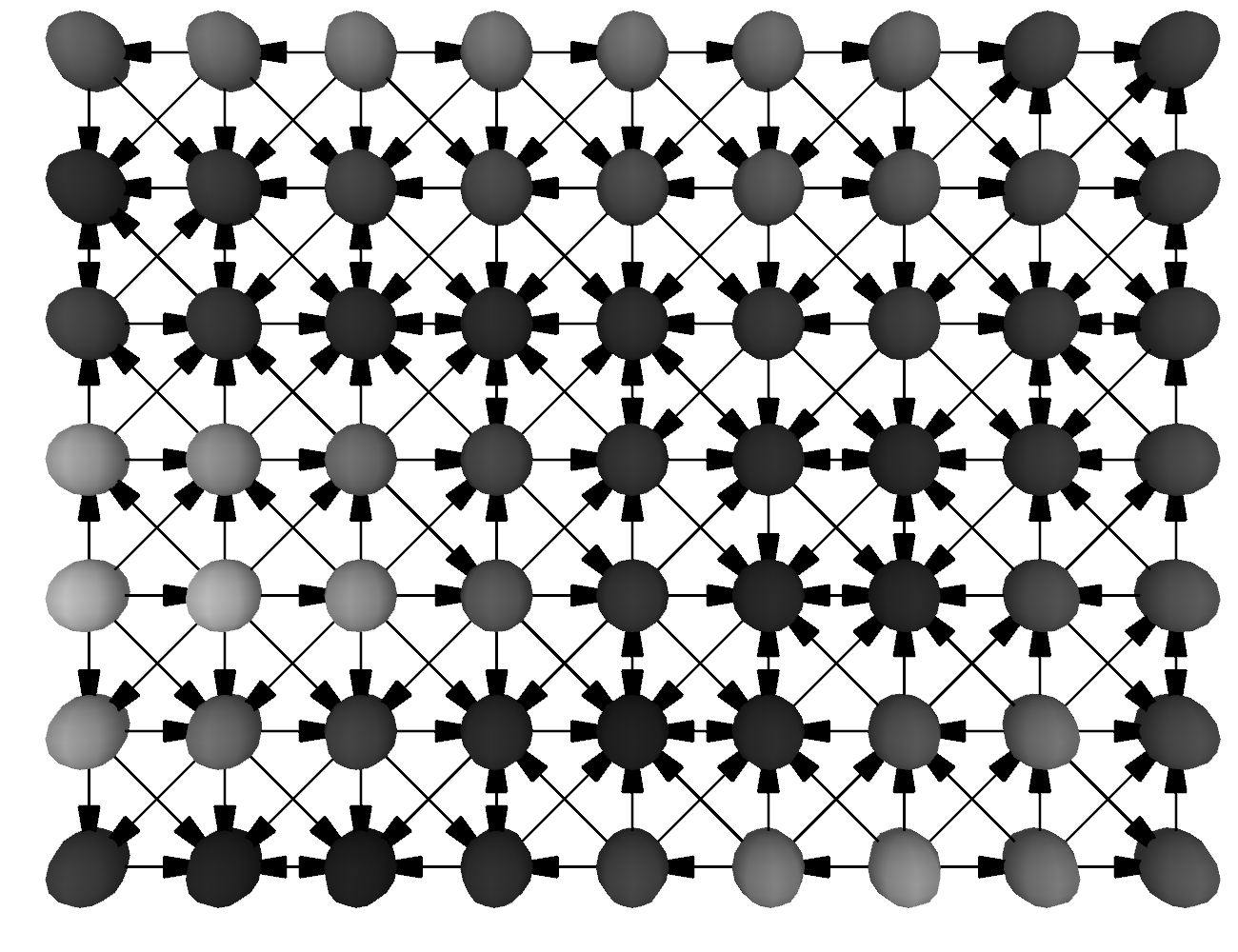}  \ \ \ \ \ \includegraphics[width=0.22\textwidth]{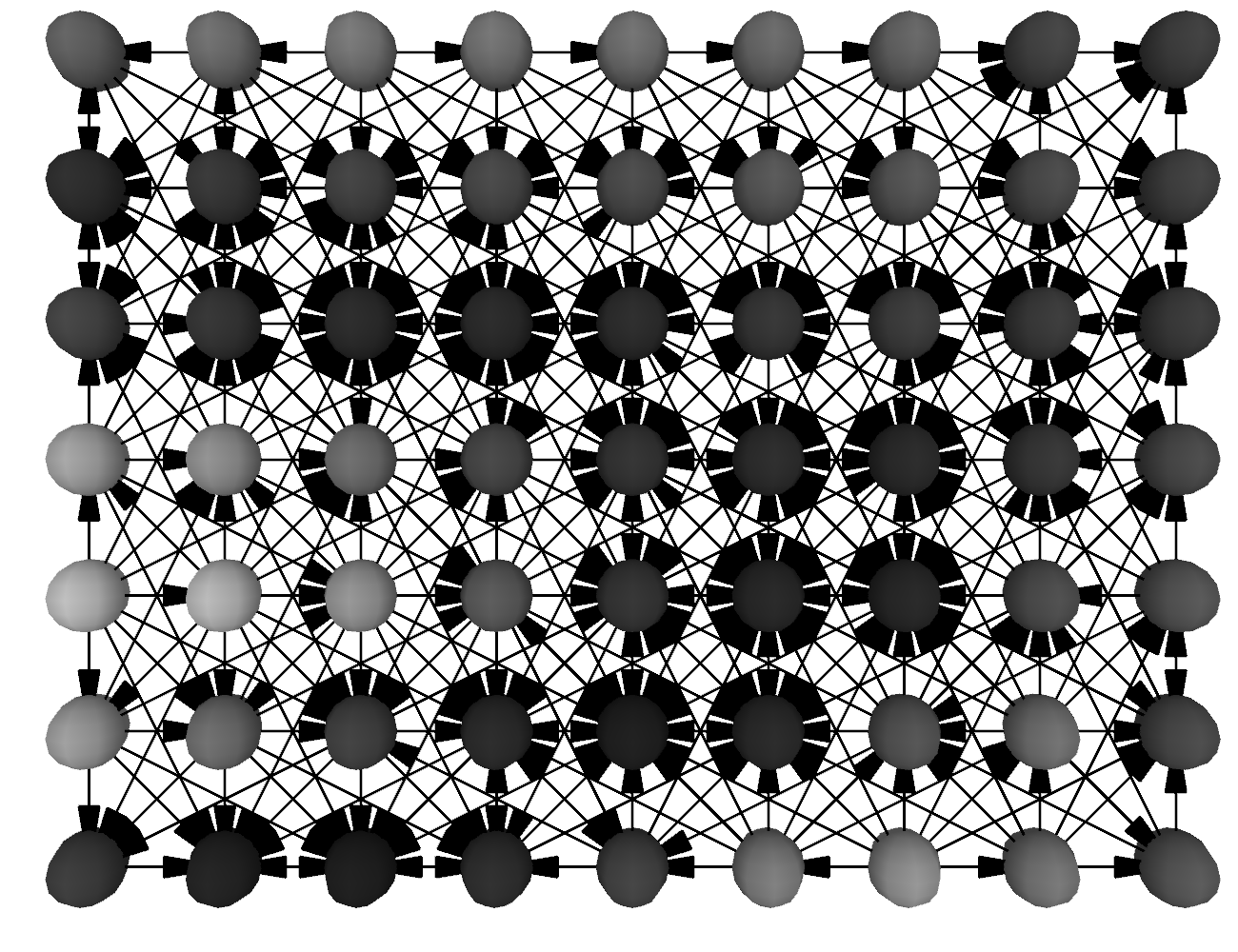} }\\
   
     \subfigure[Visual representation of network measures.]{\includegraphics[width=0.7\textwidth]{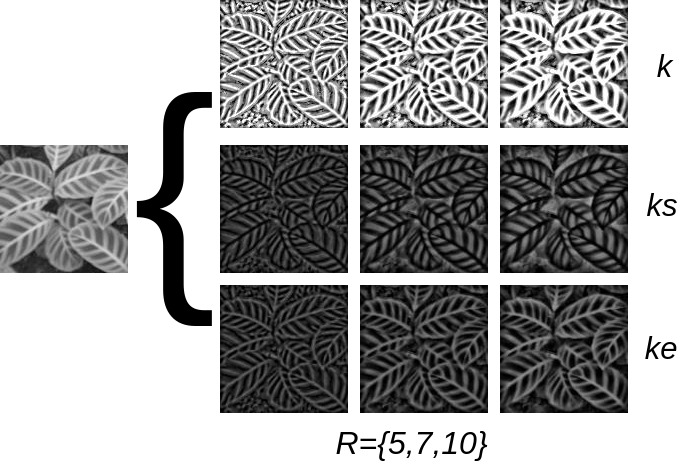}}
    
	\caption{The structure of directed complex networks (a) and a visual representation of different vertex measures obtained from networks modeled with the input image.}
	\label{fig:model_dir_gray}
\end{figure*}

The connection weight plays an important role in the network structure, and it is crucial for computing the vertex strength. It is defined as a combination between the intensity difference and the spatial position of the pixels

\begin{equation} 
 e_{v_i,v_j} = 
\left\{\begin{matrix}
 \frac{\mid I(i) - I(j) \mid}{L}, & \textnormal{if } r=1  \\ 
\frac{ \left( \frac{d(v_i,v_j)-1}{r-1} \right) + \left(\frac{\mid I(i) - I(j) \mid}{L} \right)}{2}, & \textnormal{otherwise.}
\end{matrix}\right.
\end{equation}
where $r=1$ is the smallest possible distance, thus we consider only the intensity information. This equation yields an edge distribution normalized in the range $[0,1]$, giving equal weight to both the intensity and the spatial position of the pixels. By varying the radius parameter $r$, it is possible to control the network density, i.e. as $r$ increases, the number of connections increases. With a set of radii $R\{r_1,...,r_n\}$, different networks $N^1, N^2,..., N^n$ can be modeled where one can analyze the dynamic evolution of the system (image), which is related to the interplay between neighbour pixels and, therefore, contains multiscale texture information. Figure \ref{fig:model_dir_gray}(a) illustrates the structure of the directed network modeled for different values of $r$.

To quantify the network structure, three centrality measures are employed just as in \cite{ribas2018fusion}, which are the output degree, input and output weighted degree (or strength). The out-degree $k_{v_i}$ of a vertex $v_i$ is computed by Eq. \ref{eq:out_degree} and counts the number of vertices that $v_i$ points. On the other hand, the weighted out-degree $ks_{v_i}$ considers the weight of each connection leaving $v_i$.
The weighted in-degree $ke_{v_i}$ is then the opposite of the out-degree, i.e. it sums the edges that points to $v_i$,

\begin{equation}
ke_{v_i} = \sum_{\forall v_j \in V}^{} \left\{\begin{array}{rcl}
	e_{v_j,v_i} ,& e_{v_j,v_i}\in E \\	
    0,& \textnormal{otherwise}.
	\end{array}\right.
\end{equation}.

Figure \ref{fig:model_dir_gray}(b) shows a visual representation of the network measures computed for a given image texture for different radii, where pixels are obtained normalizing each measure by the maximum possible vertex degree. It is possible to notice that each topological measure represents a different local pattern related to the image intensity variation. The combination of these measures in a multiscale fashion provides rich texture information that we exploit to train neural networks to produce image descriptors. More details are given in the following section.

\subsection{Learning Local Complex Features}

The proposal of this paper is to use the randomized neural network to learn the main topological characteristics of a CN and then use the weights of the output layer of the trained RNN as a signature to represent the CN (ie. the texture).
To achieve this, for each vertex of the CN, three information were considered: out-degree, weighted out-degree and weighted in-degree. As the out-degree is directly related to the in-degree in the modeled networks (i.e., the sum of the two degrees produces the same value in all vertices) and therefore have the same information, only the out-degree was considered.

\begin{figure}[!h]
 \centering
 \includegraphics[scale=0.7]{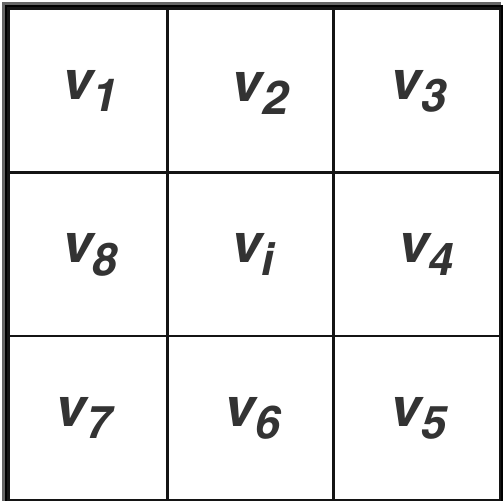}
 \caption{Neighborhood of a vertex.}
  \label{fig:Window}
\end{figure}

In this method, we propose to apply windows of size
 $3 \times 3$ over the modeled network in order to build
 the matrix of input feature vectors for the randomized neural network.
For this, firstly, consider that the Cartesian coordinates $ x_i $ and $ y_i $ of a vertex $ v_i $ are the same of the pixel $ i $ that is represented by the vertex.
Thus, the window is the spatial neighborhood of a vertex based on the Cartesian coordinates.
In other words, we divided the image into $3 \times 3$ joint windows, however, instead of using information directly from the pixels we use the information of the vertices that represent them (ie. from the CN).
Figure \ref{fig:Window} illustrates a window with a central pixel $i$ represented by the vertex $v_i$ and neighboring pixels represented by their vertices.
Regarding this window, for each vertex $v_i$, the out-degree $k_{v_i}$ is considered the label and the out-degrees of the neighboring vertices compose the input feature vector $\vec{x}_{v_i} = [k_{v_1},k_{v_2},k_{v_3}, ..., k_{v_8}]$.
This strategy to define the input feature vectors and the labels to train the neural network is the main difference between this approach and the previous work \cite{ribas2018fusion}.
In the proposed approach, it is possible to compute the feature vector using a unique value of radius $R$ for modeling, while in the previous work is necessary to use a set of values of radius to obtain a feature vector, increasing the computational time.
In addition, in the proposed approach, instead of using the gray-level we use the out-degree as the label.
Thus, the input feature vector and label are composed only of information from the CN.

\begin{figure*}[!h]
 \centering
 \includegraphics[scale=0.27]{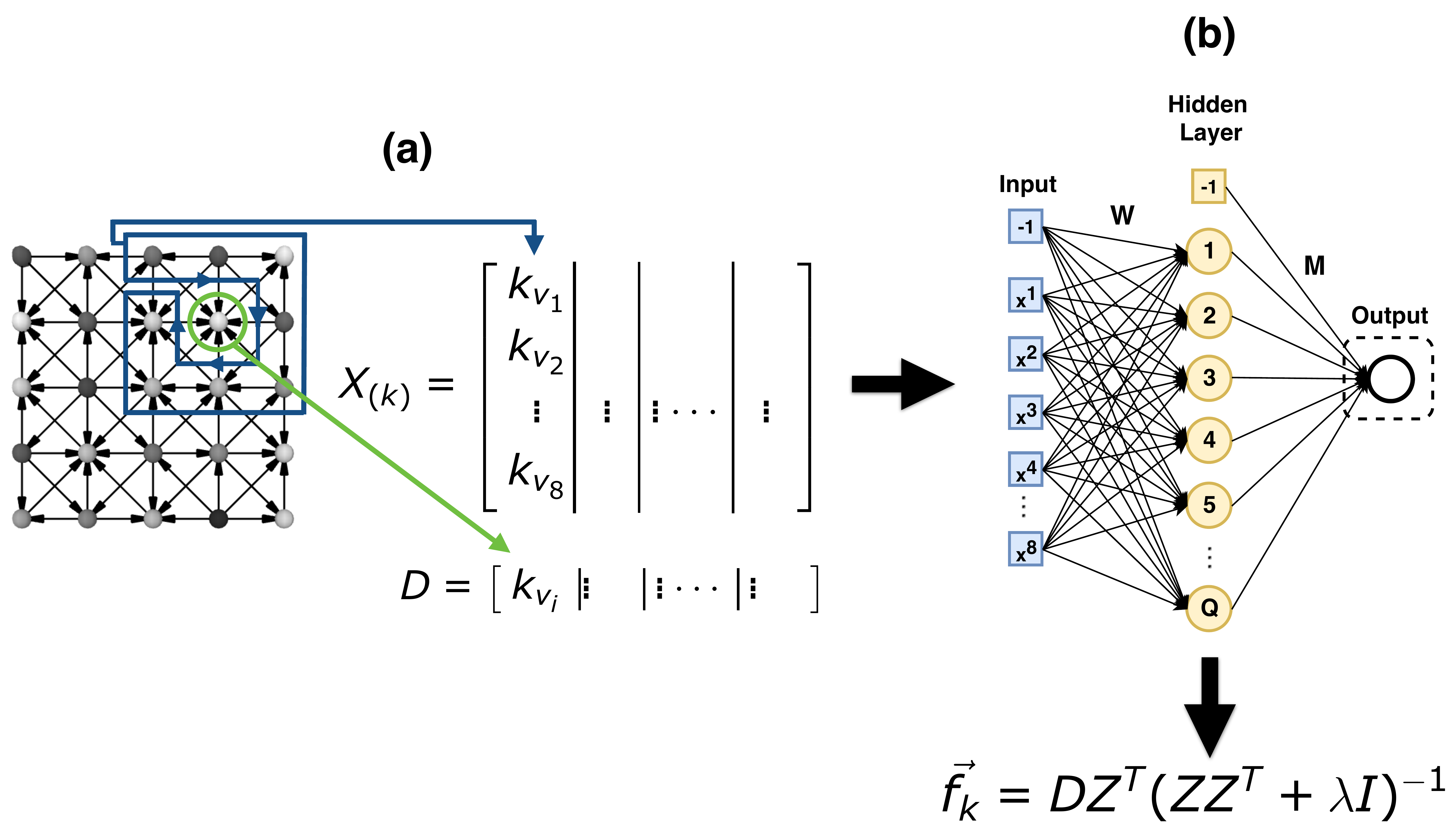}
 \caption{Label and input vector obtained from a network.}
 \label{fig:XandD}
\end{figure*}

Figure \ref{fig:XandD}(a) shows how the input feature vector and its respective label are obtained from a given vertex using a window $3\times3$.
A matrix of input feature vectors $X_{(k)}$ and a matrix of labels $D$ for the out-degree are then constructed considering all the vertices of the network.
Thus, it is possible to analyze the characteristics of the vertices that represent the pixels.
In addition to the matrix of input feature vector $X_{(k)}$ containing the out-degree, we also construct matrices of input vectors using the weighted out-degree $X_{(ks)}$ and the weighted in-degree  $X_{(ke)}$.

To train a randomized neural network, it is necessary to define the matrix of weights $W$ of the hidden neurons.
Generally, this matrix is randomly defined at each new training, changing also the values of the trained output weights.
However, in feature extraction techniques it is important
that the signature values be always equal for the same sample.
Therefore, it is important to use the same values in the matrix of weighs $W$ to obtain the same signature for the same network.
In this way, we used the same procedure adopted in \cite{JarbasRNN2015} to obtain the pseudorandom uniform numbers for the matrix $W$, that is, the linear congruent generator (LCG) \cite{park1988random}, according to

\begin{equation}
V(n+1) = (a*V(n)+b) \textnormal{ mod } c,
\end{equation}
where $V$ is the random sequence and the values $a$, $b$ and $c$ are parameters set up as $a=E + 2$, $b = E + 3$ e $c = E^2$ (values adopted in \cite{JarbasRNN2015}), where $E = Q * (p + 1)$ is the length of the sequence $V$ that is started by $V(1) = E + 1$.
Thus, the matrix $W$ is defined using the sequence $V$ divided into $Q$ segments of values $p+1$. The matrices $W$ and $X$ (each row) are normalized to zero mean and unit variation.

The descriptors of the modeled CN and, consequently, the signature of the texture is obtained based on the matrix $M$, which is composed of the learned weights of the output layer. This one becomes a vector $\vec{f} = DZ^T(ZZ^T+\lambda I)^{-1}$, where $\lambda=10^{-3}$, as illustrated in Figure \ref{fig:XandD}(b). Notice that $\vec{f}$ has a length $Q+1$ due to the bias value. Thus, the first signature for a texture is obtained by concatenating the vectors $\vec{f}$ of three RNNs trained with the three matrices of input feature vectors
$X_{(k)}$,$X_{(k)}$,$X_{(k)}$,

\begin{equation}
\vec{\Upsilon}(Q)_r = \left[ \vec{f_{k}}, \vec{f_{ks}}, \vec{f_{ke}} \right],
\end{equation}
where $Q$ is the number of hidden neurons and $r$ is the radius to construct the CN.
The signature $\vec{\Upsilon}(Q)_r$ is constructed using an unique value of $Q$ and $r$. These two parameters influence the trained weights of the neural network and, therefore, provide different descriptors for different values. Taking advantage of this, we propose a signature $\vec{\Theta}(R)_{(Q_1, Q_2, Q_m)}$, which concatenates the vectors $\vec{\Upsilon}(Q)_r$ for different values of radius $r$, 

\begin{equation}
\vec{\Theta}(Q)_{r_1, r_2, ..., r_m} = \left[ \vec{\Upsilon}(Q)_{r_1}, \vec{\Upsilon}(Q)_{r_2}, ..., \vec{\Upsilon}(Q)_{r_m} \right].
\end{equation}

Finally, a signature $\vec{\Psi}_{Q_1, Q_2, ..., Q_n}$ which combines the vector $\vec{\Theta}(Q)_{r_1, r_2, ..., r_m}$ using different values of $Q$ is proposed, according to

\begin{equation}
 \scriptscriptstyle
\vec{\Psi}_{Q_1, Q_2, ..., Q_n} = \left[ \vec{\Theta}(Q_1)_{r_1, r_2, ..., r_m}, \vec{\Theta}(Q_2)_{r_1, r_2, ..., r_m}, ..., \vec{\Theta}(Q_n)_{r_1, r_2, ..., r_m} \right].
\end{equation}


\section{Experimental Setup} \label{sec:experiments}

To assess the performance of the proposed method, the following databases were evaluated:

\begin{itemize}
	
	\item Brodatz \cite{brodatz-1966}: just as in \cite{cndt}, we used 111 classes, 16 images of 128 $\times$ 128 pixels per class, totaling a data set of 1776 images.
	
	\item Vistex \cite{Vistex1995}: this data set is provided by the MIT Media Laboratory. Just as in \cite{cndt}, we used 54 classes, each one represented by an image 512 $\times$ 512 pixels divided into non-overlapping images of 128 $\times$ 128 pixels. Thus, the database has 16 images per class, thus resulting in a total of 864 images.
	
	\item Outex \cite{OjalaMPVKH02}: this framework provides several texture benchmark data sets, and, among them, we chose TC\_Outex\_00013, which is composed of 68 classes. Just as in \cite{cndt}, each class is represented by an image of 746 $\times$ 538 pixels, from which 20 images of 128 $\times$ 128 pixels were cropped without overlapping. Thus, the database used in this work has 1360 images.
	
	\item USPTex \cite{backes2012}: this database is composed of 191 classes, 12 images per class, resulting in 2292 images, which represent a wide variety of scenes of everyday life, such as bark, sand, bricks, vegetation, sidewalk etc.  

\end{itemize}

All color texture data sets were converted into grayscale. To classify the proposed signature, we used Linear Discriminant Analysis (LDA) \cite{Webb2002}. This statistical method aims to project the data in order to maximize the distance inter-classes and minimize the variance within the same class. To validate the classification procedure, we used the \emph{leave-one-out cross-validation} approach, which basically separates one sample for test, uses the remainder for training, and repeats this process $N$ times ($N$ is the number of samples), each time with a different sample for test. The validation performance is the mean accuracy of the $N$ runs. 

\section{Results and Discussion} \label{sec:results}

\subsection{Parameter Analysis}

\begin{figure}
 \centering
 \includegraphics[scale=0.45]{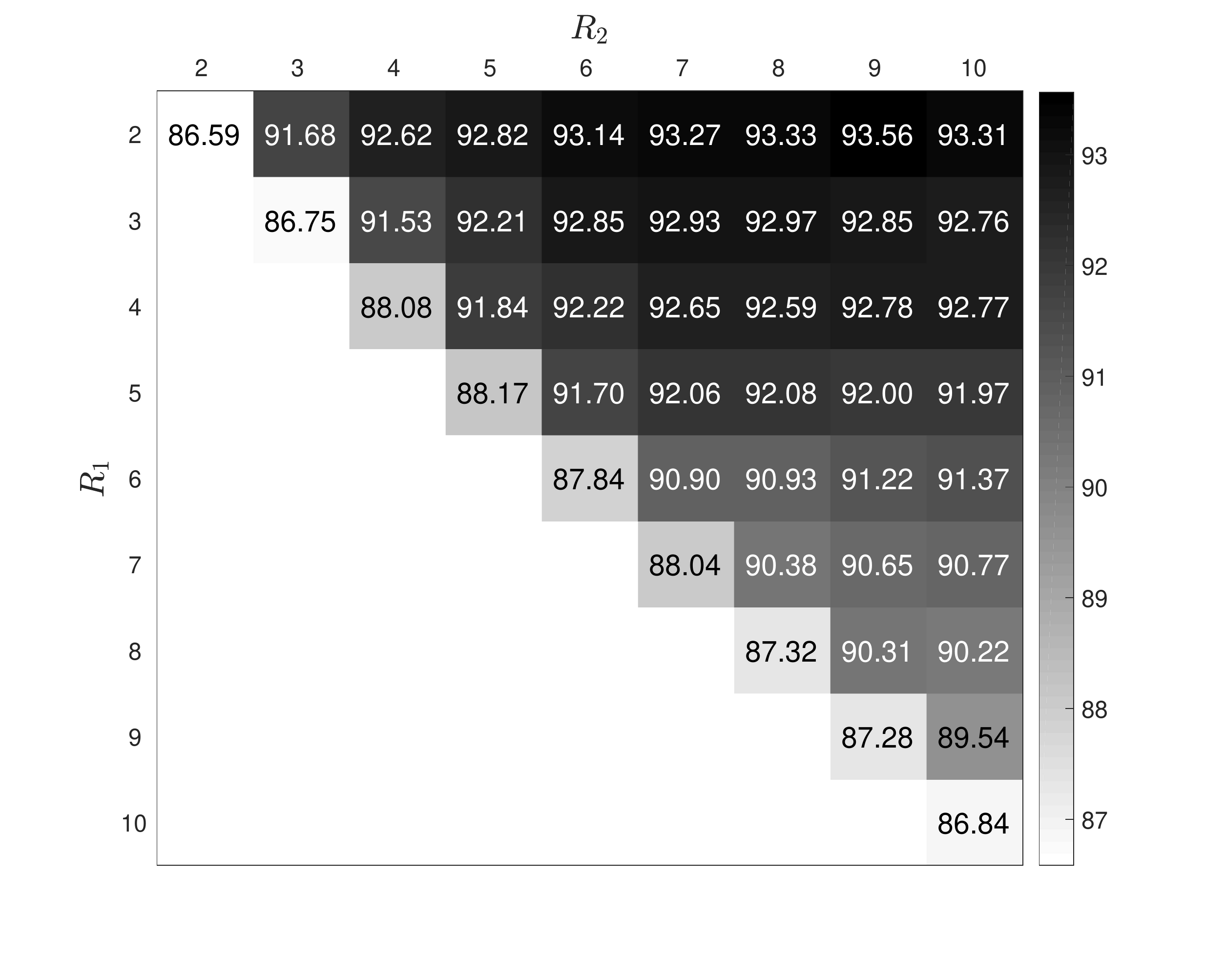}
 \caption{Accuracies of the signature $\vec{\Theta}(Q)_{R_1, R_2, ..., R_m}$ using $Q=4$ for different values of $R$ on the four databases.}
 \label{fig:valuesR}
\end{figure}

In this section, we perform an evaluation of the parameters of our method in terms of accuracy on the four databases. The first signature analyzed is the
$\vec{\Theta}(Q)_{R_1, R_2, ..., R_m}$ with $Q=4$. For this signature, different values of $R = \{2,3,4,5,6,7,8,9,10\}$ and their combinations were analyzed.
The results are summarized in Figure \ref{fig:valuesR}, which presents the mean of the accuracies obtained in the four databases. In this experiment, a maximum of two values of $R$ was combined due to the large number of characteristics generated for larger combinations.

The results indicate that for a single value of $R$ (main diagonal of the matrix), the best result is reached with $R=5$ (88.17\%). However, it is clear that a single value of $R$ yields results lower than that of the combination of two values of $R$. In this case, note that the best results are obtained when combining a low $R$ value with a high $R$ value (best result of 93.56\% for $R=\{2,9\}$. The values of $R$ determine the radius of the network modeling, that is, for low $R$ values the pixels are connected with nearest neighbors (local analysis), whereas for high $R$ values, a global analysis of texture is performed. Thus, the combination of local and global characteristics is expected to provide superior results.

\begin{figure*}
 \centering
  \subfigure[Outex]{\includegraphics[scale=0.32]{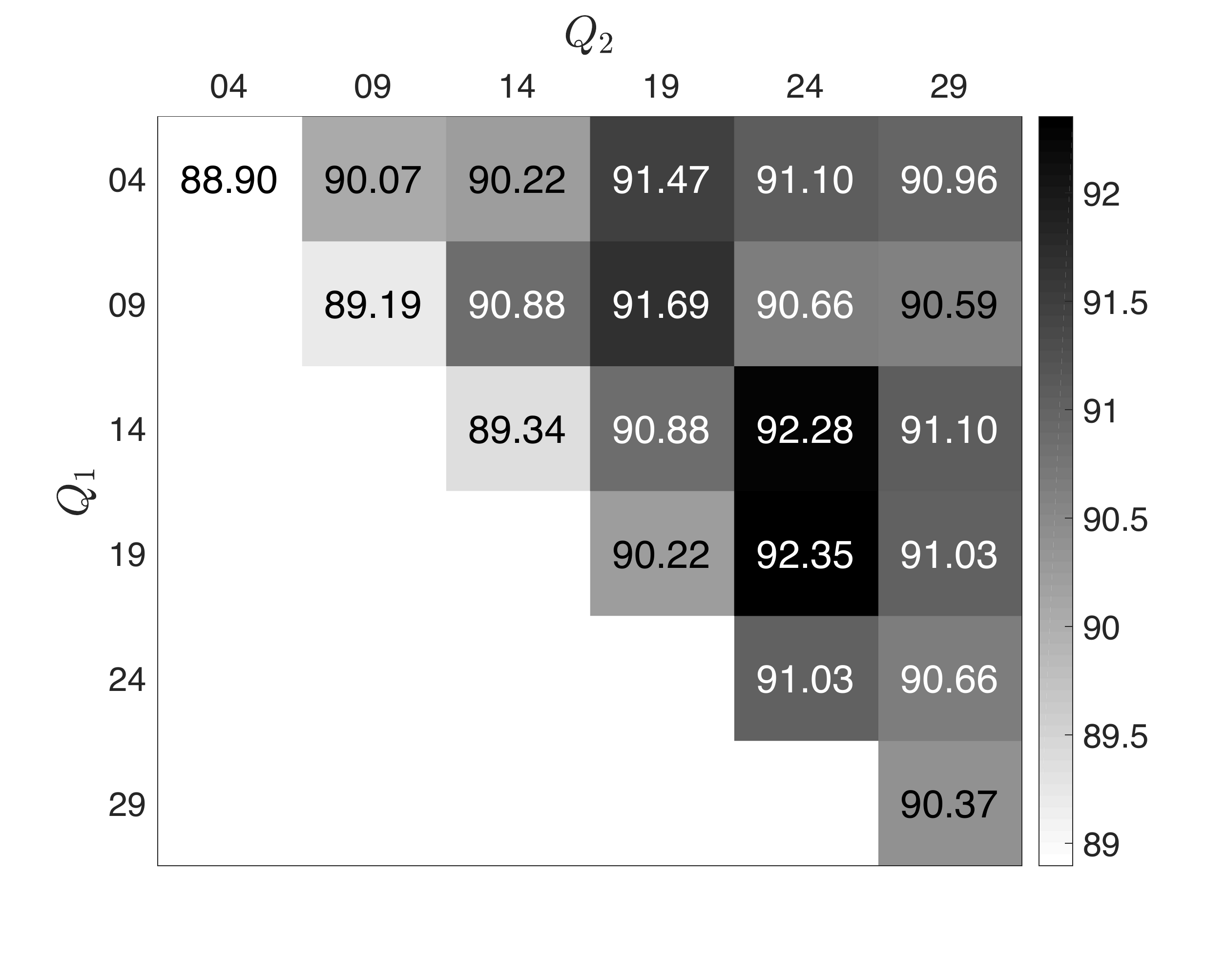}}
  \subfigure[Brodatz]{\includegraphics[scale=0.32]{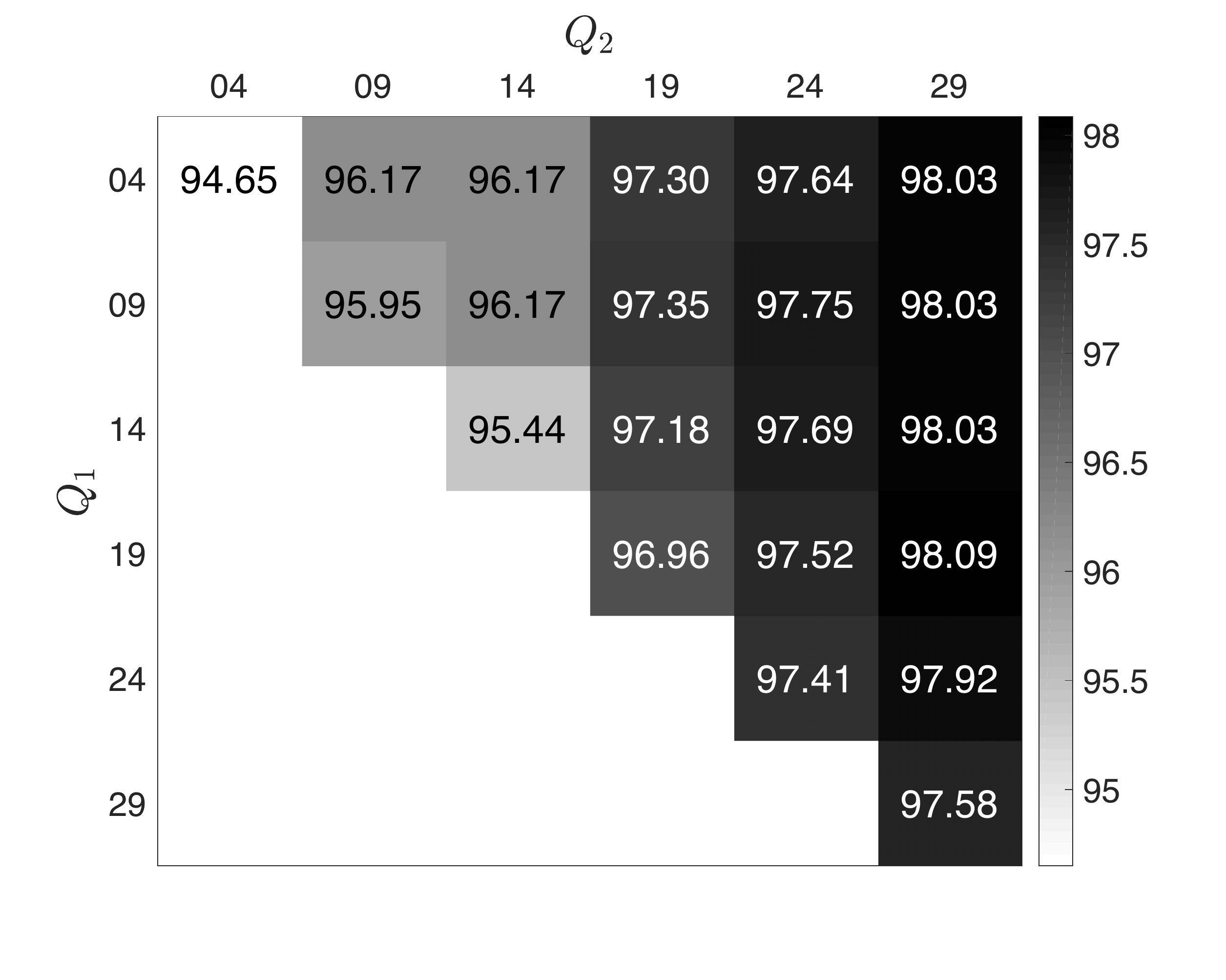}}\\
  \subfigure[USPTex]{\includegraphics[scale=0.32]{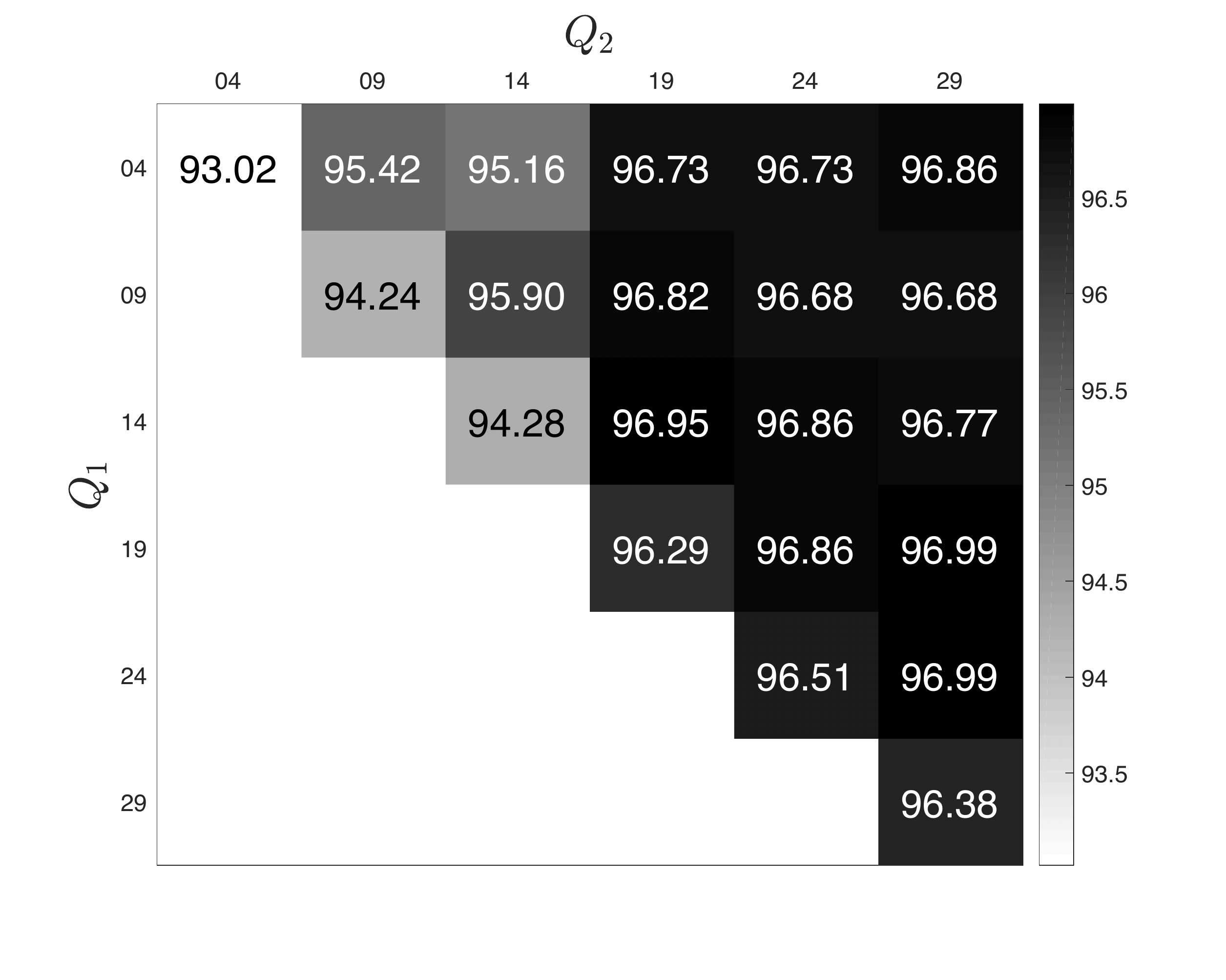}}
  \subfigure[Vistex]{\includegraphics[scale=0.32]{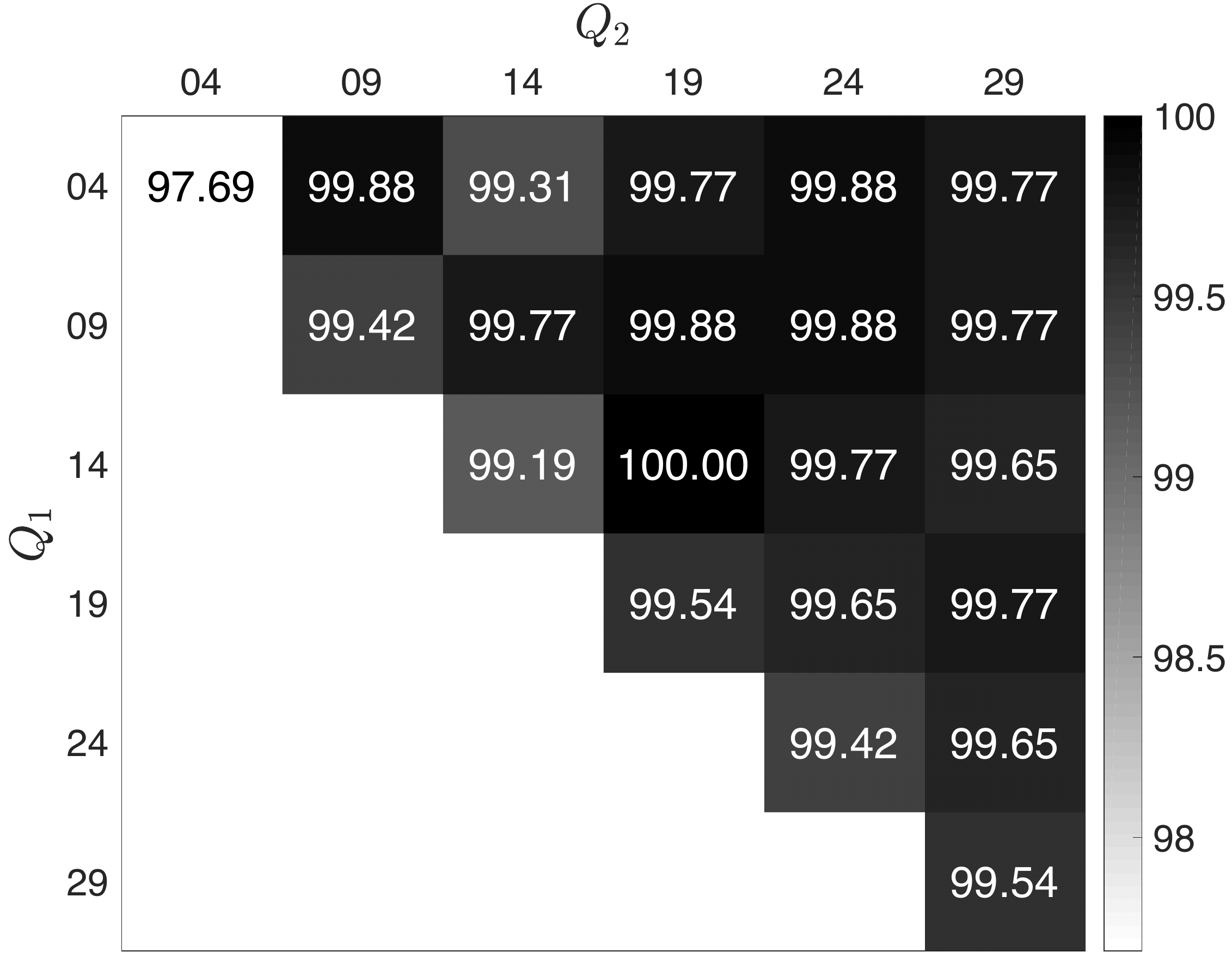}}
  \caption{Accuracies of the feature vector $\vec{\Psi}_{Q_1, Q_2, ..., Q_n}$ using different values of $Q$ on the four databases.}
 \label{fig:valuesQ}
\end{figure*}

A second experiment to evaluate the feature vector $\vec{\Psi}_{Q_1, Q_2, ..., Q_n}$ was performed. In this experiment, different values of $Q = \{ 04,09,14,19,24,29\}$ were used for $r=\{2,9\}$. Figure \ref{fig:valuesQ} shows the accuracies with the signature $\vec{\Psi}_{Q_1, Q_2, ..., Q_n}$ on the four databases with different values of $Q$ and their combinations. Note that the lower results are on the main diagonals, that is, when a single value of $Q$ is considered. On the other hand, when two values of $Q$ are considered, the accuracy increases. Table \ref{table:3Q} shows the results for the combination of three values of $Q$. Note that higher accuracies were obtained with these combinations. However, as larger values of $Q$ are combined, the size of the feature vector also increases and the accuracy tends to decrease or stabilize. Thus, we consider the vector $\vec{\Psi}_{04,19,29}$, which presents a good balance between accuracy and number of features on the four databases.

\begin{table} 
\caption{Results for the feature vector $\vec{\Psi}_{Q_1, Q_2, ..., Q_n}$ combining three values of $Q$.}
\
\centering   
\resizebox{\linewidth}{!}{
\begin{tabular}{cccccc} \hline
$Q$ & No of features & Outex & Brodatz & USPTex & Vistex \\ \hline
\{04, 09, 14\} & 180 & 91.69 & 96.45 & 96.29 & 99.77 \\     
\{04, 09, 19\} & 210 & 91.76 & 97.41 & 97.03 & 100.00 \\    
\{04, 09, 24\} & 240 & 91.25 & 97.75 & 96.77 & 99.88 \\     
\{04, 09, 29\} & 270 & 91.10 & 98.20 & 97.12 & 99.88 \\     
\{04, 14, 19\} & 240 & 91.76 & 97.52 & 96.90 & 99.88 \\     
\{04, 14, 24\} & 270 & 91.99 & 98.03 & 96.95 & 99.77 \\     
\{04, 14, 29\} & 300 & 91.10 & 98.37 & 96.86 & 99.77 \\     
\{04, 19, 24\} & 300 & 92.43 & 97.97 & 97.08 & 99.77 \\     
\textbf{\{04, 19, 29\}} & \textbf{330} & \textbf{92.13} & \textbf{98.09} & \textbf{97.21} & \textbf{99.88} \\     
\{04, 24, 29\} & 360 & 90.96 & 97.97 & 96.99 & 99.65 \\     
\{09, 14, 19\} & 270 & 91.91 & 97.47 & 97.16 & 100.00 \\    
\{09, 14, 24\} & 300 & 91.25 & 97.80 & 97.21 & 99.77 \\     
\{09, 14, 29\} & 330 & 91.10 & 98.31 & 97.21 & 100.00 \\    
\{09, 19, 24\} & 330 & 91.47 & 97.92 & 97.16 & 99.77 \\     
\{09, 19, 29\} & 360 & 91.84 & 98.25 & 97.21 & 99.88 \\     
\{09, 24, 29\} & 390 & 90.74 & 97.97 & 97.03 & 99.77 \\     
\{14, 19, 24\} & 360 & 91.91 & 97.92 & 97.25 & 99.77 \\    
\{14, 19, 29\} & 390 & 91.99 & 98.03 & 97.29 & 99.88 \\    
\{14, 24, 29\} & 420 & 91.18 & 97.92 & 97.21 & 99.77 \\    
\{19, 24, 29\} & 450 & 91.10 & 97.86 & 97.03 & 99.54 \\ \hline
\end{tabular}     
}
\label{table:3Q}                          
\end{table}

\subsection{Comparison with other methods} 

In order to evaluate the results obtained by the proposed method, comparisons were performed with other methods of the literature. The experimental configurations were the same as described above, with the exception of the CLBP method, which used the classifier 1-Nearest Neighborhood (1-NN) with the distance chi-square according to the original paper. For the proposed method, the signature that obtained the best result in the previous analysis was considered: $\vec{\Psi}_{04,19,29}$.

We also compare the performance of our descriptor with deep convolutional neural networks, which are applied in a transfer-learning approach where a pre-trained architecture is used as feature extractor by computing the Global Average Pooling (GAP) \cite{lin2013network} over the output of its last convolutional layer. The weights of these models are learned from the ImageNet dataset \cite{deng2009imagenet}, composed of millions of images, and can be ported to various applications, such as texture analysis \cite{cimpoi2016deep}. We considered four well-known models from the literature, the VGG19 (2014) \cite{simonyan2014very}, InceptionV3 (2016) \cite{szegedy2016rethinking}, ResNet50 (2016) \cite{he2016deep} and InceptionResNetV2 (2017) \cite{szegedy2017inception}. All models and their pre-trained weights were imported from the Keras 2.2.4 library \footnote{\url{www.keras.io/applications}}. For these methods, the images are normalized by the maximum possible gray-level before being processed. 

Table \ref{tab:comp_textures} presents the results obtained by our method and the others on the four texture databases. The results show that the proposed method obtained the best results when compared to the others on three databases (Outex, USPTex and Vistex). In relation to the methods based on convolutional neural networks, the ResNet50 model outperformed the proposed method only on the Brodatz database. On this database, all other approaches based on convolutional neural networks achieved higher results. However, it is possible to note that these methods did not have a good performance on the Outex database, which is characterized by samples with different illuminations.

It is also important to highlight that the proposed method improved the accuracy when compared to the ELM Signature and CNDT method (which are based on neural network nets and complex networks, respectively). The ELM Signature method uses only the pixel intensities to train the RNNs and characterize the textures, which shows that modeling in CNs and their topology are important in texture analysis. On the other hand, the CNDT method models the images in complex networks and calculates traditional measures of CN to compose the feature vector. In this context, the results suggest that our approach of learning complex network characteristics using randomized neural networks to characterize textures is more discriminative.

\begin{table}[!h]  
\centering  
\caption{Comparison of accuracies of different texture analysis methods in four texture databases.}  
\resizebox{\linewidth}{!}{
\begin{tabular}{lrcccc} \hline                               
Methods & No of features & Outex & USPTex & Brodatz & Vistex \\
\hline
GLCM \cite{haralick1979statistical}  &  24 & 80.73 & 83.64 & 90.43 & 92.24 \\ 
GLDM \cite{weszka1976comparative}  & 60 & 86.76 & 92.06 & 94.43 & 97.11 \\  
Gabor Filters \cite{manjunath1996texture}  & 48 & 81.91 &  89.22 & 89.86 & 93.29 \\ 
Fourier \cite{weszka1976comparative}  & 63 & 81.91 & 67.50 & 75.90 & 79.51 \\    
Fractal \cite{backes2009plant}   & 69 & 80.51 & 78.27 & 87.16 & 91.67 \\ 
Fractal Fourier \cite{florindo2012fractal}   & 68 & 68.38 & 59.47 & 71.96 & 79.75 \\ 
LBP \cite{ojala2002multiresolution} & 256 & 81.10 & 85.43 & 93.64 & 97.92 \\                      
LBPV \cite{lbpv}  & 555 & 75.66 & 54.97 & 86.26 & 88.65 \\
CLBP \cite{clbp} & 648 & 85.80 & 91.14 & 95.32 & 98.03 \\                       
AHP \cite{zhu2015adaptive} & 120 & 88.31 & 94.85 & 94.88 & 98.38 \\    
BSIF \cite{kannala2012bsif}  & 256 & 77.43 & 77.66 & 91.44 & 88.66 \\                     
LCP \cite{lcp}  & 81 & 86.25 & 91.14 & 93.47 & 94.44 \\                       
LFD \cite{lfd} & 276 & 82.57 & 83.55 & 90.99 & 94.68 \\                      
LPQ \cite{lpq} & 256 & 79.41 & 85.12 & 92.51 & 92.48 \\                      
ELM Signature \cite{JarbasRNN2015}  & 180 & 89.71 & 95.11 & 95.27 & 98.15 \\ 
CNTD \cite{cndt}  & 108  & 86.76 & 91.71 & 95.27 & 98.03 \\
CNRNN\_1 \cite{ribas2018fusion} & 180 & 91.54 & 96.64 & 96.11 & 98.73 \\  
CNRNN\_2 \cite{ribas2018fusion} & 240 & 91.32 & 96.95 & 96.06 & 99.19 \\ \hline
VGG19~\cite{simonyan2014very} & 512 & 76.62 & 93.19 & 96.79 & 97.45\\
InceptionV3~\cite{szegedy2016rethinking} & 2048 & 86.40 & 96.77 & 98.54 & 98.84 \\
ResNet50~\cite{he2016deep} & 2048 & 65.66 & 62.30 & 81.98 & 81.71 \\
InceptionResNetV2~\cite{szegedy2017inception} & 1536 & 85.88 & 96.34 & \textbf{98.99} & 98.96 \\
\hline  
Proposed approach & 330  & \textbf{92.13} & \textbf{97.21} & 98.09 & \textbf{99.88}
\\ \hline        
\end{tabular}  
}
\label{tab:comp_textures}                  
\end{table} 

\section{Conclusion} \label{sec:conclusion}

This paper proposed a novel way of extracting information from complex networks to train randomized neural networks in order to use the weights of output neurons to compose a texture signature. Unlike the method proposed in \cite{ribas2018fusion}, our approach proposed a different strategy to construct the input feature vector and the label, which is based on CN information only. This one uses a unique value of radius to model the complex network and train the neural network, decreasing the computational time. The success rates of our approach were very high, surpassing accuracies of a large set of compared methods, including some based on deep learning. Thus, in the light of the obtained results, we believe that our approach offers a relevant contribution to the novel and promising field of research that studies how to connect neural and complex networks to build image signatures.

\section*{Acknowledgment}

 Lucas Correia Ribas gratefully acknowledges the financial support grant \#s 2016/23763-8 and 16/18809-9, S\~ao Paulo Research Foundation (FAPESP). Jarbas Joaci de Mesquita S\'a Junior thanks CNPq (National Council for Scientific and Technological Development, Brazil) (Grant: 302183/2017-5) for the financial support of this work.  Leonardo Scabini acknowledges support from CNPq (Grant number \#142438/2018-9). Odemir M. Bruno thanks the financial support of CNPq (Grant \# 307897/2018-4) and FAPESP (Grant \#s 14/08026-1 and 16/18809-9).
 The authors are also grateful to the NVIDIA GPU Grant Program for the donation of the Quadro P6000 and the Titan Xp GPUs used on this research.

\bibliographystyle{acm}  
\bibliography{references.bib}  

\begin{thebibliography}{10}

\bibitem{backes2009plant}
{\sc Backes, A.~R., Casanova, D., and Bruno, O.~M.}
\newblock Plant leaf identification based on volumetric fractal dimension.
\newblock {\em International Journal of Pattern Recognition and Artificial
  Intelligence 23}, 06 (2009), 1145--1160.

\bibitem{backes2012}
{\sc Backes, A.~R., Casanova, D., and Bruno, O.~M.}
\newblock Color texture analysis based on fractal descriptors.
\newblock {\em Pattern Recognition 45}, 5 (2012), 1984--1992.
\newblock {h}ttp://scg.ifsc.usp.br/dataset/USPtex.php.

\bibitem{cndt}
{\sc Backes, A.~R., Casanova, D., and Bruno, O.~M.}
\newblock Texture analysis and classification: A complex network-based
  approach.
\newblock {\em Information Sciences 219\/} (2013), 168--180.

\bibitem{barabasi2016network}
{\sc Barab{\'a}si, A., and P\'osfai, M.}
\newblock {\em Network Science}.
\newblock Cambridge University Press, 2016.

\bibitem{scalefreeCN}
{\sc Barab{\'a}si, A.-L., and Albert, R.}
\newblock Emergence of scaling in random networks.
\newblock {\em science 286}, 5439 (1999), 509--512.

\bibitem{basu2016theoretical}
{\sc Basu, S., Karki, M., Mukhopadhyay, S., Ganguly, S., Nemani, R., DiBiano,
  R., and Gayaka, S.}
\newblock A theoretical analysis of deep neural networks for texture
  classification.
\newblock In {\em 2016 International Joint Conference on Neural Networks
  (IJCNN)\/} (2016), IEEE, pp.~992--999.

\bibitem{brodatz-1966}
{\sc Brodatz, P.}
\newblock {\em Textures: {A} photographic album for artists and designers}.
\newblock Dover Publications, New York, 1966.

\bibitem{calvetti2000}
{\sc Calvetti, D., Morigi, S., Reichel, L., and Sgallari, F.}
\newblock Tikhonov regularization and the {L}-curve for large discrete
  ill-posed problems.
\newblock {\em Journal of Computational and Applied Mathematics 123}, 1 (2000),
  423 -- 446.

\bibitem{cimpoi2016deep}
{\sc Cimpoi, M., Maji, S., Kokkinos, I., and Vedaldi, A.}
\newblock Deep filter banks for texture recognition, description, and
  segmentation.
\newblock {\em International Journal of Computer Vision 118}, 1 (2016), 65--94.

\bibitem{costa2007CNsurvey}
{\sc Costa, L. d.~F., Rodrigues, F.~A., Travieso, G., and Villas~Boas, P.~R.}
\newblock Characterization of complex networks: A survey of measurements.
\newblock {\em Advances in Physics 56}, 1 (2007), 167--242.

\bibitem{2004densesift}
{\sc Csurka, G., Dance, C., Fan, L., Willamowski, J., and Bray, C.}
\newblock Visual categorization with bags of keypoints.
\newblock In {\em ECCV International Workshop on Statistical Learning in
  Computer Vision\/} (2004), pp.~1--22.

\bibitem{JarbasRNN2015}
{\sc de~Mesquita~{Sá Junior}, J.~J., and Backes, A.~R.}
\newblock Elm based signature for texture classification.
\newblock {\em Pattern Recognition 51\/} (2016), 395 -- 401.

\bibitem{DEVES20142925}
{\sc de~Ves, E., Acevedo, D., Ruedin, A., and Benavent, X.}
\newblock A statistical model for magnitudes and angles of wavelet frame
  coefficients and its application to texture retrieval.
\newblock {\em Pattern Recognition 47}, 9 (2014), 2925 -- 2939.

\bibitem{deng2009imagenet}
{\sc Deng, J., Dong, W., Socher, R., Li, L.-J., Li, K., and Fei-Fei, L.}
\newblock Imagenet: A large-scale hierarchical image database.
\newblock In {\em Computer Vision and Pattern Recognition, 2009. CVPR 2009.
  IEEE Conference on\/} (2009), Ieee, pp.~248--255.

\bibitem{randomCN}
{\sc Erdos, P., and R{\'e}nyi, A.}
\newblock On random graphs i.
\newblock {\em Publ. Math. Debrecen 6\/} (1959), 290--297.

\bibitem{randomCNevolution}
{\sc Erdos, P., and R{\'e}nyi, A.}
\newblock On the evolution of random graphs.
\newblock {\em Publ. Math. Inst. Hungar. Acad. Sci 5\/} (1960), 17--61.

\bibitem{florindo2012fractal}
{\sc Florindo, J.~B., and Bruno, O.~M.}
\newblock Fractal descriptors based on {F}ourier spectrum applied to texture
  analysis.
\newblock {\em Physica A: statistical Mechanics and its Applications 391}, 20
  (2012), 4909--4922.

\bibitem{gonccalves2016texture}
{\sc Gon{\c{c}}alves, W.~N., da~Silva, N.~R., da~Fontoura~Costa, L., and Bruno,
  O.~M.}
\newblock Texture recognition based on diffusion in networks.
\newblock {\em Information Sciences 364\/} (2016), 51--71.

\bibitem{lcp}
{\sc Guo, Y., Zhao, G., and Pietik{\"a}inen, M.}
\newblock Texture classification using a linear configuration model based
  descriptor.
\newblock In {\em BMVC\/} (2011), Citeseer, pp.~1--10.

\bibitem{clbp}
{\sc Guo, Z., Zhang, L., and Zhang, D.}
\newblock A completed modeling of local binary pattern operator for texture
  classification.
\newblock {\em IEEE Transactions on Image Processing 19}, 6 (2010), 1657--1663.

\bibitem{lbpv}
{\sc Guo, Z., Zhang, L., and Zhang, D.}
\newblock Rotation invariant texture classification using lbp variance (lbpv)
  with global matching.
\newblock {\em Pattern recognition 43}, 3 (2010), 706--719.

\bibitem{haralick1979statistical}
{\sc Haralick, R.~M.}
\newblock Statistical and structural approaches to texture.
\newblock {\em Proceedings of the IEEE 67}, 5 (1979), 786--804.

\bibitem{haralick1973}
{\sc Haralick, R.~M., Shanmugam, K., and Dinstein, I.~H.}
\newblock Textural features for image classification.
\newblock {\em Systems, Man and Cybernetics, IEEE Transactions on}, 6 (1973),
  610--621.

\bibitem{he2016deep}
{\sc He, K., Zhang, X., Ren, S., and Sun, J.}
\newblock Deep residual learning for image recognition.
\newblock In {\em Proceedings of the IEEE conference on computer vision and
  pattern recognition\/} (2016), pp.~770--778.

\bibitem{huang2006extreme}
{\sc Huang, G.-B., Zhu, Q.-Y., and Siew, C.-K.}
\newblock Extreme learning machine: theory and applications.
\newblock {\em Neurocomputing 70}, 1 (2006), 489--501.

\bibitem{kannala2012bsif}
{\sc Kannala, J., and Rahtu, E.}
\newblock Bsif: Binarized statistical image features.
\newblock In {\em Pattern Recognition (ICPR), 2012 21st International
  Conference on\/} (2012), IEEE, pp.~1363--1366.

\bibitem{lam1997rotated}
{\sc Lam, W.-K., and Li, C.-K.}
\newblock Rotated texture classification by improved iterative morphological
  decomposition.
\newblock {\em IEE Proceedings-Vision, Image and Signal Processing 144}, 3
  (1997), 171--179.

\bibitem{lazebnik2005sparse}
{\sc Lazebnik, S., Schmid, C., and Ponce, J.}
\newblock A sparse texture representation using local affine regions.
\newblock {\em IEEE Transactions on Pattern Analysis and Machine Intelligence
  27}, 8 (2005), 1265--1278.

\bibitem{lin2013network}
{\sc Lin, M., Chen, Q., and Yan, S.}
\newblock Network in network.
\newblock {\em arXiv preprint arXiv:1312.4400\/} (2013).

\bibitem{lfd}
{\sc Maani, R., Kalra, S., and Yang, Y.-H.}
\newblock Noise robust rotation invariant features for texture classification.
\newblock {\em Pattern Recognition 46}, 8 (2013), 2103--2116.

\bibitem{manjunath1996texture}
{\sc Manjunath, B.~S., and Ma, W.-Y.}
\newblock Texture features for browsing and retrieval of image data.
\newblock {\em IEEE Transactions on pattern analysis and machine intelligence
  18}, 8 (1996), 837--842.

\bibitem{Moore1920}
{\sc Moore, E.~H.}
\newblock On the reciprocal of the general algebraic matrix.
\newblock {\em Bulletin of the American Mathematical Society 26\/} (1920),
  394--395.

\bibitem{OjalaMPVKH02}
{\sc Ojala, T., M{\"a}enp{\"a}{\"a}, T., Pietik{\"a}inen, M., Viertola, J.,
  Kyll{\"o}nen, J., and Huovinen, S.}
\newblock Outex: New framework for empirical evaluation of texture analysis
  algorithms.
\newblock In {\em International Conference on Pattern Recognition\/} (2002),
  pp.~701--706.

\bibitem{ojala2002multiresolution}
{\sc Ojala, T., Pietikainen, M., and Maenpaa, T.}
\newblock Multiresolution gray-scale and rotation invariant texture
  classification with local binary patterns.
\newblock {\em IEEE Transactions on pattern analysis and machine intelligence
  24}, 7 (2002), 971--987.

\bibitem{lpq}
{\sc Ojansivu, V., and Heikkil{\"a}, J.}
\newblock Blur insensitive texture classification using local phase
  quantization.
\newblock In {\em International conference on image and signal processing\/}
  (2008), Springer, pp.~236--243.

\bibitem{panjwani1995markov}
{\sc Panjwani, D.~K., and Healey, G.}
\newblock Markov random field models for unsupervised segmentation of textured
  color images.
\newblock {\em IEEE Transactions on pattern analysis and machine intelligence
  17}, 10 (1995), 939--954.

\bibitem{pao1994learning}
{\sc Pao, Y.-H., Park, G.-H., and Sobajic, D.~J.}
\newblock Learning and generalization characteristics of the random vector
  functional-link net.
\newblock {\em Neurocomputing 6}, 2 (1994), 163--180.

\bibitem{pao1992functional}
{\sc Pao, Y.-H., and Takefuji, Y.}
\newblock Functional-link net computing: theory, system architecture, and
  functionalities.
\newblock {\em Computer 25}, 5 (1992), 76--79.

\bibitem{park1988random}
{\sc Park, S.~K., and Miller, K.~W.}
\newblock Random number generators: good ones are hard to find.
\newblock {\em Communications of the ACM 31}, 10 (1988), 1192--1201.

\bibitem{penrose_1955}
{\sc Penrose, R.}
\newblock A generalized inverse for matrices.
\newblock {\em Mathematical Proceedings of the Cambridge Philosophical Society
  51}, 3 (1955), 406–--413.

\bibitem{Vistex1995}
{\sc Picard, R., Graczyk, C., Mann, S., Wachman, J., Picard, L., and Campbell,
  L.}
\newblock {\em Vision texture database}.
\newblock Media Laboratory, MIT, Cambridge, Massachusetts, 1995.

\bibitem{ribas2018fusion}
{\sc Ribas, L.~C., de~Mesquita [Sá~Junior], J.~J., Scabini, L.~F., and Bruno,
  O.~M.}
\newblock Fusion of complex networks and randomized neural networks for texture
  analysis.
\newblock {\em Pattern Recognition 103\/} (2020), 107189.

\bibitem{Ribas2015}
{\sc Ribas, L.~C., Gon{\c{c}}alves, D.~N., Oru{\^{e}}, J. P.~M., and
  Gon{\c{c}}alves, W.~N.}
\newblock {Fractal dimension of maximum response filters applied to texture
  analysis}.
\newblock {\em Pattern Recognition Letters 65\/} (2015), 116--123.

\bibitem{SaJrRNNColor2019}
{\sc S{\'a}~Junior, J. J.~M., Backes, A.~R., and Bruno, O.~M.}
\newblock Randomized neural network based signature for color texture
  classification.
\newblock {\em Multidimensional Systems and Signal Processing 30}, 3 (2019),
  1171--1186.

\bibitem{schmidt1992feedforward}
{\sc Schmidt, W.~F., Kraaijveld, M.~A., and Duin, R. P.~W.}
\newblock Feedforward neural networks with random weights.
\newblock In {\em Proceedings., 11th IAPR International Conference on Pattern
  Recognition. Vol.II. Conference B: Pattern Recognition Methodology and
  Systems\/} (1992), pp.~1--4.

\bibitem{simonyan2014very}
{\sc Simonyan, K., and Zisserman, A.}
\newblock Very deep convolutional networks for large-scale image recognition.
\newblock {\em arXiv preprint arXiv:1409.1556\/} (2014).

\bibitem{szegedy2017inception}
{\sc Szegedy, C., Ioffe, S., Vanhoucke, V., and Alemi, A.~A.}
\newblock Inception-v4, inception-resnet and the impact of residual connections
  on learning.
\newblock In {\em Thirty-First AAAI Conference on Artificial Intelligence\/}
  (2017).

\bibitem{szegedy2016rethinking}
{\sc Szegedy, C., Vanhoucke, V., Ioffe, S., Shlens, J., and Wojna, Z.}
\newblock Rethinking the inception architecture for computer vision.
\newblock In {\em The IEEE Conference on Computer Vision and Pattern
  Recognition (CVPR)\/} (June 2016).

\bibitem{tikhonov1963}
{\sc Tikhonov, A.~N.}
\newblock On the solution of ill-posed problems and the method of
  regularization.
\newblock {\em Dokl. Akad. Nauk USSR 151}, 3 (1963), 501–--504.

\bibitem{smallworldCN}
{\sc Watts, D.~J., and Strogatz, S.~H.}
\newblock Collective dynamics of ‘small-world’networks.
\newblock {\em nature 393}, 6684 (1998), 440--442.

\bibitem{Webb2002}
{\sc Webb, A.}
\newblock {\em Statistical Pattern Recognition}, 2nd~ed.
\newblock John Wiley \& Sons Ltd, Chichester, England, 2002.

\bibitem{weszka1976comparative}
{\sc Weszka, J.~S., Dyer, C.~R., and Rosenfeld, A.}
\newblock A comparative study of texture measures for terrain classification.
\newblock {\em IEEE transactions on Systems, Man, and Cybernetics}, 4 (1976),
  269--285.

\bibitem{zhu2015adaptive}
{\sc Zhu, Z., You, X., Chen, C.~P., Tao, D., Ou, W., Jiang, X., and Zou, J.}
\newblock An adaptive hybrid pattern for noise-robust texture analysis.
\newblock {\em Pattern Recognition 48}, 8 (2015), 2592--2608.

\end{thebibliography}


\end{document}